# Did you take the pill? - Detecting Personal Intake of Medicine from Twitter

**DEBANJAN MAHATA,** *Bloomberg*
**JASPER FRIEDRICHS,** *Mountain View*
**RAJIV RATN SHAH,** *Indraprastha Institute of Information Technology Delhi*
**JING JIANG,** *Singapore Management University*

Mining social media messages such as tweets, blogs, and Facebook posts for health and drug related information has received significant interest in pharmacovigilance research. Social media sites (*e.g.*, Twitter), have been used for monitoring drug abuse, adverse reactions of drug usage and analyzing expression of sentiments related to drugs. Most of these studies are based on aggregated results from a large population rather than specific sets of individuals. In order to conduct studies at an individual level or specific cohorts, identifying posts mentioning intake of medicine by the user is necessary. Towards this objective we develop a classifier for identifying mentions of personal intake of medicine in tweets. We train a stacked ensemble of shallow convolutional neural network (CNN) models on an annotated dataset. We use random search for tuning the hyper-parameters of the CNN models and present an ensemble of best models for the prediction task. Our system produces state-of-the-art result, with a micro-averaged F-score of 0.693. We believe that the developed classifier has direct uses in the areas of psychology, health informatics, pharmacovigilance and affective computing for tracking moods, emotions and sentiments of patients expressing intake of medicine in social media.

## 1. INTRODUCTION

Social media has become a ubiquitous source of information for various topics. Right from information related to daily events, personal rants, to expressions of intake of medicine and adverse drug reactions, are readily available in publicly accessible social media channels such as Twitter[1], DailyStrength[2], MedHelp[3], among others. Huge amounts of data made available on these platforms have become a useful resource for conducting public health monitoring and surveillance, commonly known as pharmacovigilance [1]. The work presented in this paper aims at identifying intake of personal medication expressed by a user on Twitter. The broader perspective of such a system is to aid in developing automated methods for performing pharmacovigilance activities in social media, and to study the effects of medicine on an individual as well as specific cohorts [2]. Such a system would further aid in studying psychology of individuals as well as groups by tracking sentiments, emotions and moods expressed in social media after intake of a particular medicine.

Attempts have been made to mine social media content in order to identify adverse drug reactions [3], abuse [4], and user sentiment [5], from posts mentioning medications. However, all these studies are based on aggregated results from large set of content that mentions a medicine/drug, without taking into account whether the user has actually consumed the medicine/drug. Without this knowledge, a true assessment of the effects of medication intake in general and how it affects a specific group of users cannot be done. In order to leverage social media data for performing such assessments and studying targeted groups, it is necessary to develop systems that can automatically distinguish posts that expresses personal intake of medicine from those that do not. Moreover, since multimodal information has shown a great importance in sentiment analysis [6], [7], we would like to leverage them in future to effectively identify adverse drug reactions.

In this work we concentrate on Twitter as the social media channel. The key to the process of identifying tweets mentioning personal intake of medicine and to draw insights from them is to build accurate text classification systems. The effectiveness of developing classifiers has already been shown to be useful in identifying adverse drug reactions expressed in Twitter [3]. However, mining social media posts comes with unique challenges. Microblogging websites like Twitter pose challenges for automated information mining tools and techniques due to their brevity, noisiness, idiosyncratic language, unusual structure and ambiguous representation of discourse. Information extraction tasks using state-of-the-art natural language processing techniques, often give poor results for tweets. Abundance of link farms, unwanted promotional posts, and nepotistic relationships between content creates additional challenges [9].

[1]http://twitter.com
[2]https://dailystrength.org
[3]http://medhelp.org

The main objective of the task presented in this paper is to categorize short colloquial tweets into one of the following three classes,

1) **personal medication intake** (Class 1) - tweets in which the user clearly expresses a personal medication intake/consumption (*e.g.*, *I had the worst headache ever and I just took an AdvilRelief #advil and now I feel so much better thank*).
2) **possible medication intake** (Class 2) - tweets that are ambiguous but suggest that the user may have taken the medication (*e.g.*, *I should have taken advil on friday then i might have actluly had an amazing weekend. instead of throwing up 20 times a day #advil, not this time*)
3) **non-intake** (Class 3) - tweets that mention medication names but do not indicate personal intake (*e.g.*, *Understand the causes and managing #Migraine Madness #aspirin #diet #botox #advil #relpax #headache*).

Towards the above goal, we design and implement a deep learning classifier - Stacked Ensemble of Shallow Convolutional Neural Networks (see Section 2.1), trained on an annotated dataset provided at SMM4H-2017 shared task workshop[4]. We compare the results of our classification system with other classifiers that participate in the shared task and get state-of-the-art results, with a micro-averaged F-score of 0.693 for Classes 1 and 2. We submitted our system (InfyNLP) at the workshop and were ranked first amongst 26 submissions [10], [11]. In this paper, we intend to elaborately discuss and present our submitted system as well as our model choices and learning.

## 2. METHODOLOGY

Deep learning systems have recently shown to achieve top results in tasks related to natural language processing on tweets [11]. Historically, ensemble learning has proved to be very effective in most of the machine learning tasks including the famous winning solution of the Netflix Prize [12]. Ensemble models can offer diversity over model architectures, training data splits or random initialization of the same model or model architectures. Multiple average or low performing learners are combined to produce a robust and high-performing learning model.

A convolutional neural network (CNN) is a deep learning architecture that has shown strong performance on sentencelevel text classification [13]. Convolutional neural networks are effective at document classification primarily because they are able to pick out salient features (*e.g.*, tokens or sequences of tokens) in a way that is invariant to their position within the input sequence of words. Even fairly simple CNNs evaluate at a level of or even better than more complex deep learning architectures [14]. Therefore, we design and implement a stacked ensemble of shallow convolutional neural networks (see Figure 1) for solving the classification task presented in this paper. The main intuition behind developing such an ensemble was to take the best of all worlds. Next, we explain the architecture of stacked ensemble of CNNs that we train.

**2.1 Stacked Ensemble of Shallow Convolutional Neural Networks (CNNs)**

A stacked ensemble of shallow CNNs is a large ensemble classifier comprising of smaller ensembles stacked over one another, prioritized by their performance, with the underlying classifier being a standard shallow CNN model similar to that used in the work [13]. In order to train such an ensemble model we enlist the generic steps:

---

[4]https://healthlanguageprocessing.org/sharedtask2/

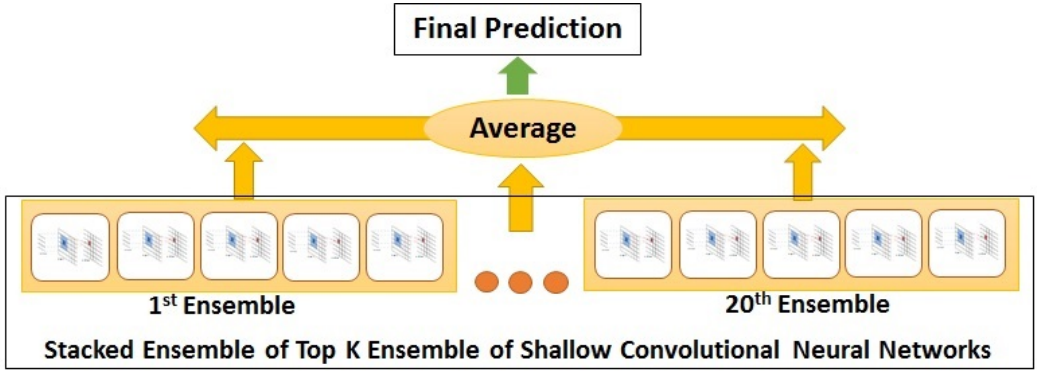

**Figure 1:** A stacked ensemble of 100 (20 x 5) shallow convolutional neural networks.

**Step 1.** Train a shallow CNN model on each fold while performing *c*-fold cross validation on the training dataset.
**Step 2.** The output prediction of each model trained on each fold is averaged to get the final prediction of an ensemble of *c* CNN models (see Equation 1).
**Step 3.** Train *n* such ensembles as in Step 2.
**Step 4.** Sort the *n* ensembles in terms of their performance on the metric suitable for the classification task.
**Step 5.** Choose top *K* ensembles based on their performance on the training dataset to form the final stacked ensemble of *K* CNN ensemble models.
**Step 6.** The final output prediction is given by the average of the predictions made by each of the top *K* ensembles (see Equation 2).

$$prediction_i^{ensemble} = \frac{1}{c}\sum_{j=1}^{c} prediction_{i_j} \quad (1)$$

$$prediction_K^{stacked} = \frac{1}{K}\sum_{k=1}^{K} prediction_{top_K}^{ensemble} \quad (2)$$

Figure 1, shows a high level architecture of the final stacked ensemble of CNNs that we use in predicting the outcome of the task presented in this paper. We train a standard shallow CNN model, on each fold while performing 5-fold cross validation on our training dataset. We take the output prediction of each of these models trained on each fold and average them to create an ensemble of 5 models. We further train 99 such ensembles. For the final prediction we sort the ensembles in order of their decreasing performance on the training dataset and take the top *K* ensembles. We take the prediction of each of the *K* ensembles and average them to get the final prediction from our stacked ensemble of shallow CNNs. In general, we can take top *K* such ensembles and create a stacked ensemble of top *K* ensemble of shallow CNNs.

In order to get the best results from any classification model, hyperparameter tuning is a key step and CNNs are no exception. While the existing literature offers guidance on practical design decisions, identifying the best hyperparameters of a CNN requires experimentation. This requires evaluating trained models on a cross-validation dataset and manually choosing the hyperparameters that produce the best results. Automated hyperparameter searching methods like grid search, random search, and Bayesian optimization methods are also commonly used. In our presented system we use random search [15], to explore the hyperparameters of a shallow CNN architecture and form an ensemble of the best models, which we refer to as a stacked ensemble. Next, we share the detailed settings, output and analysis of our experiment.

## 3. EXPERIMENT

In this section, we present our experiments. We give an overview of the dataset on which we train our models, and discuss the hyperparameter settings. Results of our experiments are presented, accompanied by a discussion of the metrics used for evaluation and comparison with other models trained on the same dataset.

**3.1 Dataset**

The dataset used in this paper is publicly available and can be obtained from the 2nd Social Media Mining for Health Applications Shared Task at AMIA 2017 website[5]. The organizers of the task provided 8000 annotated tweets as a training dataset and 2260 additional tweets as development dataset. We collected the tweets using the script provided along with the dataset, by querying Twitter's API. However, we could not collect all the tweets as some of them were not available at the moment when we executed our collection process. Later, the organizers also shared the test dataset, which was used for calculating the final scores of the submitted models. The test dataset consists of 7513 tweets. A distribution of tweets provided for each class and the mapping of each class is shown in Table 1. It is to be noted that for training our models, we combine the training and development dataset provided and treat it as our training dataset, therefore learning our models using 9663 tweets with 5-fold cross validation.

**Table 1:** Shared task data distribution. Classes 1, 2 and 3 represent personal medication intake, possible medication intake and no medication intake, respectively.

|  | Class 1 | Class 2 | Class 3 | Total |
|---|---|---|---|---|
| **Train** | 1847 | 3027 | 4789 | 9663 |
| **Test** | 1731 | 2697 | 3085 | 7513 |

**3.2 Data Preprocessing and CNN Architecture**

We use Spacy[6] for all our data preprocessing and cleaning activities. We do not remove stopwords. Each document in our training and test dataset is converted to a fixed size document of 47 words/tokens. We use two pre-trained word embeddings - godin [16] and shin [17], shared by the authors. Each of these embeddings are of 400 dimensions. Each word in the input tweet is represented by its corresponding embedding vector, when present in the vocabulary of the model. Tweets are mapped to embedding vectors and are available as a matrix input to the model. Convolutions are performed across the input word-wise using differently sized filters. The resulting feature maps are then processed using a max pooling layer to condense or summarize the extracted features. The final layer consists of a fully-connected dense neural network with the extracted features as the input and a soft-max output. The final model is trained using the procedure described in Section 2 along with the choice of hyperparamters as explained next.

**Table 2**: Hyperparameter ranges used for random search permutations.

| Hyperparameter | Range |
|---|---|
| Word Embedding Model | godin [16], shin [17] |
| No. of Filters | 100, 200, 300, 400 |
| Filter Sizes | [1,2,3,4,5], [2,3,4,5,6], [3,4,5,6,7], [1,2,2,2,3], [2,3,3,3,4], [3,4,4,4,5], [4,5,5,5,6] |
| Dense Layer Size | 100, 200, 300, 400 |
| Dropout Probability | 0.4, 0.5, 0.6, 0.7, 0.8, 0.9 |
| Batch Size | 50, 100, 150 |
| Learning Rate | 0.0001, 0.001 |
| Adam beta2 | 0.9, 0.999 |

[5] https://healthlanguageprocessing.org/sharedtask2/

## 3.3 Hyperparameter Settings for CNNs

We use Xavier weight initialization scheme [18], for initializing the weights of the CNNs. Adam [19] with two annealing restarts has been shown to work faster and perform better than SGD in other NLP tasks [20]. Therefore, we use the same as our optimization algorithm. We use five filters with varying filter sizes in the convolution layer and use dropout during the training process. The models are implemented using TensorFlow[7]. The entire ranges of the hyperparameters that we give to our random search procedure is shown in Table 2. The word embedding model to be used during training is also treated as a hyperparameter.

One of the key aspects of CNNs are its filters and the choice of filters while designing the architecture. Different filter sizes allow grouping of word representations at different scales. We also explore the performance of filters of different sizes for the CNN ensembles that we train. By keeping the filter size fixed we train four runs of CNN ensembles by grid search on other hyperparameters, namely learning rate and filter size. Figure 2, shows the average performances of individual filter sizes for four runs along with their standard deviations. A filter size of 5 gives the best performance as evident from the figure. We permute over the best five such filter sizes for our hyperparameter settings.

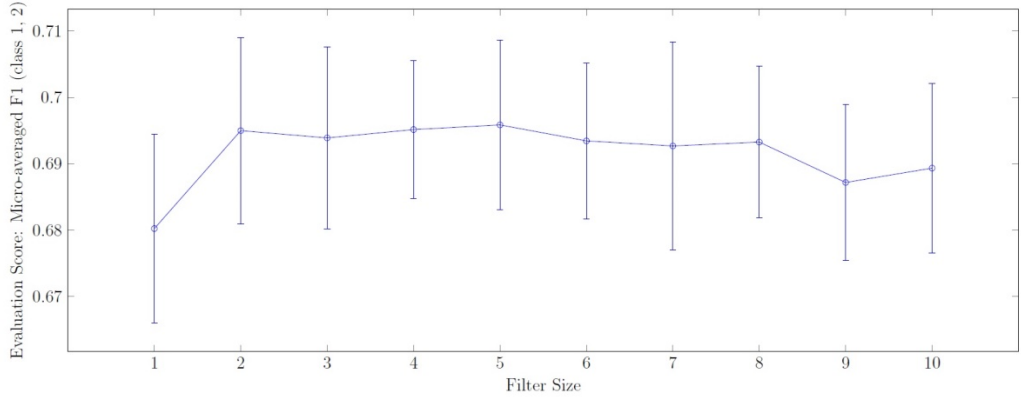

**Figure 2:** Average and standard deviation of the performances of ensembles of CNN models on four runs, with a fixed filter size.

## 3.4 Evaluation Metric

The evaluation metric used was micro-averaged F-score ($F_{1+2}$) of the class 1 (personal medication intake) and class 2 (possible medication intake), for assessing the performance of our model, as used in the *Social Media Mining for Health* shared task [10]. The equation for calculating the micro- averaged F-score for classes 1 and 2, which in turn depends on micro-averaged precision ($P_{1+2}$) and recall ($R_{1+2}$) for classes 1 and 2, is shown in equations 3, 4 and 5, respectively.

$$F_{1+2} = \frac{2 \times P_{1+2} \times R_{1+2}}{P_{1+2} + R_{1+2}} \quad (3)$$

$$P_{1+2} = \frac{TP_1 + TP_2}{TP_1 + FP_1 + TP_2 + FP_2} \quad (4)$$

$$R_{1+2} = \frac{TP_1 + TP_2}{TP_1 + FN_1 + TP_2 + FN_2} \quad (5)$$

where, $TP_i$ is the number of True Positives for *Class i*; $TN_i$ is the number of True Negatives for *Class i*; $FP_i$ is the number of False Positives for *Class i*; $FN_i$ is the number of False Negatives for *Class i*.

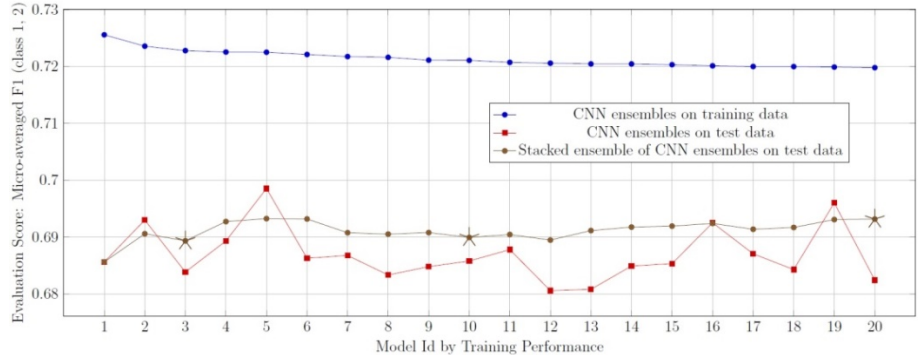

**Figure 3:** Performances for top 20 individual 5-fold CNN ensembles and collective stacked-ensemble of CNN ensembles arranged in decreasing order.

**3.5 Results and Discussion**

An ensemble of five CNNs is trained during 5-fold cross-validation training performed on our combined training dataset along with random search on the hyperparameter ranges. We train 99 such ensembles. The performance of the top 20 individual ensembles on the training data (blue) and on the test data (red) is shown in Figure 3. The models are arranged in the order of their decreasing training performances. It can be also observed from Figure 3, that the fifth best individual 5-fold CNN ensemble achieves the best scores on the test dataset. We create stacked ensembles from these ensembles by taking top *K* ensemble models. We show the performances for such top *K* stacked ensembles (brown), as well. The detailed performances on the evaluation metrics of top 3, top 10 and top 20 stacked ensembles are shown in Tables 3 and 4, and denoted by stars in Figure 3. The stacked ensemble formed using top 20 best performing ensembles achieved the best micro averaged F1 score on the test dataset. This proves an overall effectiveness of ensemble models in boosting performance on the present classification task.

We compare the performance of our system with some of the top performing systems in the SMM4H workshop [9]. These systems represent the current state-of-the-art performances on the given dataset and task. With the exception of NRC-Canada that implements a Support Vector Machine classifier using a variety of surface-form, sentiment, and domain-specific features, all the other systems attempt to solve the task using convolutional neural networks. UKNLP trained a CNN model with attention mechanism. CSaRUS-CNN uses a cost sensitive and random undersampling variants of CNNs. TurkuNLP developed an ensemble of neural networks with features generated by word and character-level convolutional neural network channels and a condensed weighted bag-of-words representation. There is a clear indication of ensembles and CNNs being the dominant strategy of choice in implementing high-performing systems for the task presented in this paper.

**Table 3:** Evaluation of stacked ensembles on test data w.r.t. precision and recall.

|  | **Precision** | | | | **Recall** | | |
|---|---|---|---|---|---|---|---|
|  | **1** | **2** | **3** | | **1** | **2** | **3** |
| **Top3** | 0.696 | 0.644 | 0.842 | | 0.704 | 0.725 | **0.763** |
| **Top10** | 0.685 | 0.646 | 0.849 | | 0.709 | 0.729 | 0.758 |
| **Top20** | **0.690** | **0.648** | **0.853** | | **0.712** | **0.733** | 0.761 |

**Table 4:** Evaluation of stacked ensembles on test data w.r.t. F1 and micro averaged scores. * marks the state-of-the-art micro averaged F1 on the task's dataset achieved by our best model.

|  | **F1** | | | $R_{1+2}$ | $P_{1+2}$ | $F_{1+2}$ |
|---|---|---|---|---|---|---|
|  | 1 | 2 | 3 | | | |
| **Top3** | 0.700 | 0.682 | 0.800 | 0.664 | 0.716 | 0.689 |
| **Top10** | 0.697 | 0.685 | 0.801 | 0.661 | 0.721 | 0.690 |
| **Top20** | 0.701 | 0.688 | 0.804 | 0.664 | 0.725 | 0.693* |

Table 5: Performance comparison of our system with the other state-of-the-art systems.

| Systems | Micro-Avg Precision Class 1 and 2 ($P_{1+2}$) | Micro-Avg Recall Class 1 and 2 ($R_{1+2}$) | Micro-Avg F-score Class 1 and 2 ($F_{1+2}$) |
|---|---|---|---|
| **Our System** | **0.725** | 0.664 | **0.693** |
| UKNLP | 0.701 | **0.677** | 0.689 |
| NRC-Canada | 0.704 | 0.635 | 0.668 |
| TurkuNLP | 0.701 | 0.630 | 0.663 |
| CSaRUS-CNN | 0.709 | 0.604 | 0.652 |

## 5. CONCLUSIONS AND FUTURE WORK

In this paper we showed the generic effectiveness of CNNs and ensembles on identification of personal medication intake from Twitter posts. Our proposed architecture of stacked ensemble of shallow CNNs, out-performed other models. This provided an empirical evaluation of our initial aim of combining ensembles with CNNs along with training the models using random search on the hyperparameters. In the future, we plan to work more on hyperparameter tuning using random search and various other search procedures and analyze their effectiveness. Instead of using pre-trained word embeddings it would also be interesting to look at the performance of our models by training word and phrase embeddings on a domain specific dataset of tweets. We would also like to use the classier for studying moods and emotions of social media users expressing intake of medicine and plan to use our system in solving some of the problems that lies at the intersection of pharmacovigilance, affective computing and psychology.